\documentclass[letterpaper, 10 pt, conference]{ieeeconf} 
\IEEEoverridecommandlockouts
\usepackage{siunitx}
\usepackage{physics}
\usepackage[T1]{fontenc}
\AtBeginDocument{\RenewCommandCopy\qty\SI}
\usepackage{amsmath,amsfonts}
\usepackage{array}
\usepackage{cite}
\usepackage[caption=false,font=scriptsize]{subfig}
\usepackage{textcomp}
\usepackage{stfloats}
\usepackage{float}
\usepackage{url}
\usepackage{verbatim}
\usepackage{graphicx}
\usepackage{amssymb}
\usepackage{pifont}
\usepackage{algorithm}
\usepackage{algpseudocode}
\usepackage[disable]{todonotes} % disabled because it breaks tikzexternalize
\usepackage{hyperref}
\usepackage[capitalize]{cleveref}
\usepackage{comment}
\usepackage{multirow}
\usepackage{booktabs}
\usepackage{rotating}
\usepackage{makecell}
\usepackage{tabularx}
\usepackage{xcolor}
\usepackage{pgf}
\usepackage{blindtext}
\usepackage{pgfplots}
\usepackage{pgfplots}
\usepgfplotslibrary{colormaps}
\pgfplotsset{compat=newest}
\pgfplotsset{colormap name=viridis} % Example colormap
\usepackage{tikz}

\def\BibTeX{{\rm B\kern-.05em{\sc i\kern-.025em b}\kern-.08em
    T\kern-.1667em\lower.7ex\hbox{E}\kern-.125emX}}

\begin{document}

\title{\LARGE \bf
GripMap: An Efficient, Spatially Resolved Constraint Framework for Offline and Online Trajectory Planning in Autonomous Racing}

\author{
Frederik Werner\thanks{F. Werner, M. Lienkamp are with the Institute of Automotive Technology, TUM School of Engineering and Design, Technical University of Munich, 85748 Garching, Germany; Munich Institute of Robotics and Machine Intelligence (MIRMI), corresponding author: frederik.werner@tum.de}, 
Ann-Kathrin Schwehn\thanks{A. Schwehn, J. Betz are with the Professorship of Autonomous Vehicle Systems, TUM School of Engineering and Design, Technical University of Munich, 85748 Garching, Germany; Munich Institute of Robotics and Machine Intelligence (MIRMI)}, 
Markus Lienkamp, Johannes Betz}

% Switch to single column mode
\onecolumn

% Manually insert the copyright notice outside of the two-column layout
\begin{center}
    \textcopyright \ 2025 IEEE. Personal use of this material is permitted. Permission from IEEE must be obtained for all other uses, including reprinting/republishing this material for advertising or promotional purposes, collecting new collected works for resale or redistribution to servers or lists, or reuse of any copyrighted component of this work in other works.
\end{center}

% Switch back to two-column mode
\twocolumn

\maketitle

%%%%%%%%%%%%%%%%%%%%%%%%%%%%%%%%%%%%%%%%%%%%%%%%%%%%%%%%%%%%%%%%%%%%%%%%%%%%%%%%

\begin{abstract}

Conventional trajectory planning approaches for autonomous vehicles often assume a fixed vehicle model that remains constant regardless of the vehicle's location. This overlooks the critical fact that the tires and the surface are the two force-transmitting partners in vehicle dynamics; while the tires stay with the vehicle, surface conditions vary with location. Recognizing these challenges, this paper presents a novel framework for spatially resolving dynamic constraints in both offline and online planning algorithms applied to autonomous racing. We introduce the GripMap concept, which provides a spatial resolution of vehicle dynamic constraints in the Frenet frame, allowing adaptation to locally varying grip conditions. This enables compensation for location-specific effects, more efficient vehicle behavior, and increased safety, unattainable with spatially invariant vehicle models. The focus is on low storage demand and quick access through perfect hashing. This framework proved advantageous in real-world applications in the presented form. Experiments inspired by autonomous racing demonstrate its effectiveness. In future work, this framework can serve as a foundational layer for developing future interpretable learning algorithms that adjust to varying grip conditions in real-time.

\end{abstract}

%%%%%%%%%%%%%%%%%%%%%%%%%%%%%%%%%%%%%%%%%%%%%%%%%%%%%%%%%%%%%%%%%%%%%%%%%%%%%%%%
\section{INTRODUCTION}

In racing, the term \textit{grip} refers to the traction between a vehicle's tires and the track surface, enabling it to accelerate, brake, and corner effectively without losing control. However, local grip can vary, which poses a significant challenge. These variations are often caused by the buildup of rubber on the racing line, which increases traction, and the accumulation of dust, tire debris, or other materials on less-used track sections, which decreases grip. Such variations can be substantial, with relative grip reductions exceeding 50\% across the racetrack surface or even laterally within the same section of the track\cite{Woodward.2012, Wadell.2019}. Additionally, external factors like rain can drastically alter the tire-road friction potential $\mu$.
Human drivers naturally account for these changes, relying on experience and visual cues to adjust their driving strategies accordingly~\cite{Werner.2024, Doubek.2021b}.

However, autonomous vehicle systems currently lack accurate, location-dependent information. Providing such information would enable trajectory planning algorithms~\cite{Teng2023} to account not only for location-specific vehicle limits due to varying grip but also for the practical constraints imposed by the software stack. Even with a well-tuned vehicle model, the system may still struggle in certain sections where the software reaches its limits. This is often due to compounding errors from modeling inaccuracies, control deviations, or localization uncertainties. As a result, the combined vehicle-control system may saturate earlier than the pure vehicle friction potential would suggest in certain regions of the track~\cite{Werner.2024, Kegelman.2018, Ganzner.2023}.

\begin{figure}
\centering
\begin{tikzpicture}[font=\scriptsize]
        \definecolor{TUMYellow}{RGB}{254, 215, 2}
        \definecolor{darkorange}{RGB}{217, 81, 23} 
        % Trackbound
        \draw [thick, TUMYellow, line width=1pt] (3.4,0) --(3.7,0); 
        \node[anchor=west] at (3.8,0) {Track bounds};
    
        % Centerline
        \draw [thick, darkorange, line width=1pt] (5.8, 0) -- (6.1,0); 
        \node[anchor=west] at (6.2,0) {Centerline};
    
         % GripMap Cells
        \draw[thick, black, line width=1pt] (7.8,0) -- (8.1,0);
        \node[anchor=west] at (8.2, 0) {GripMap Cells};
        
    \end{tikzpicture}
\includegraphics[width=1\columnwidth]{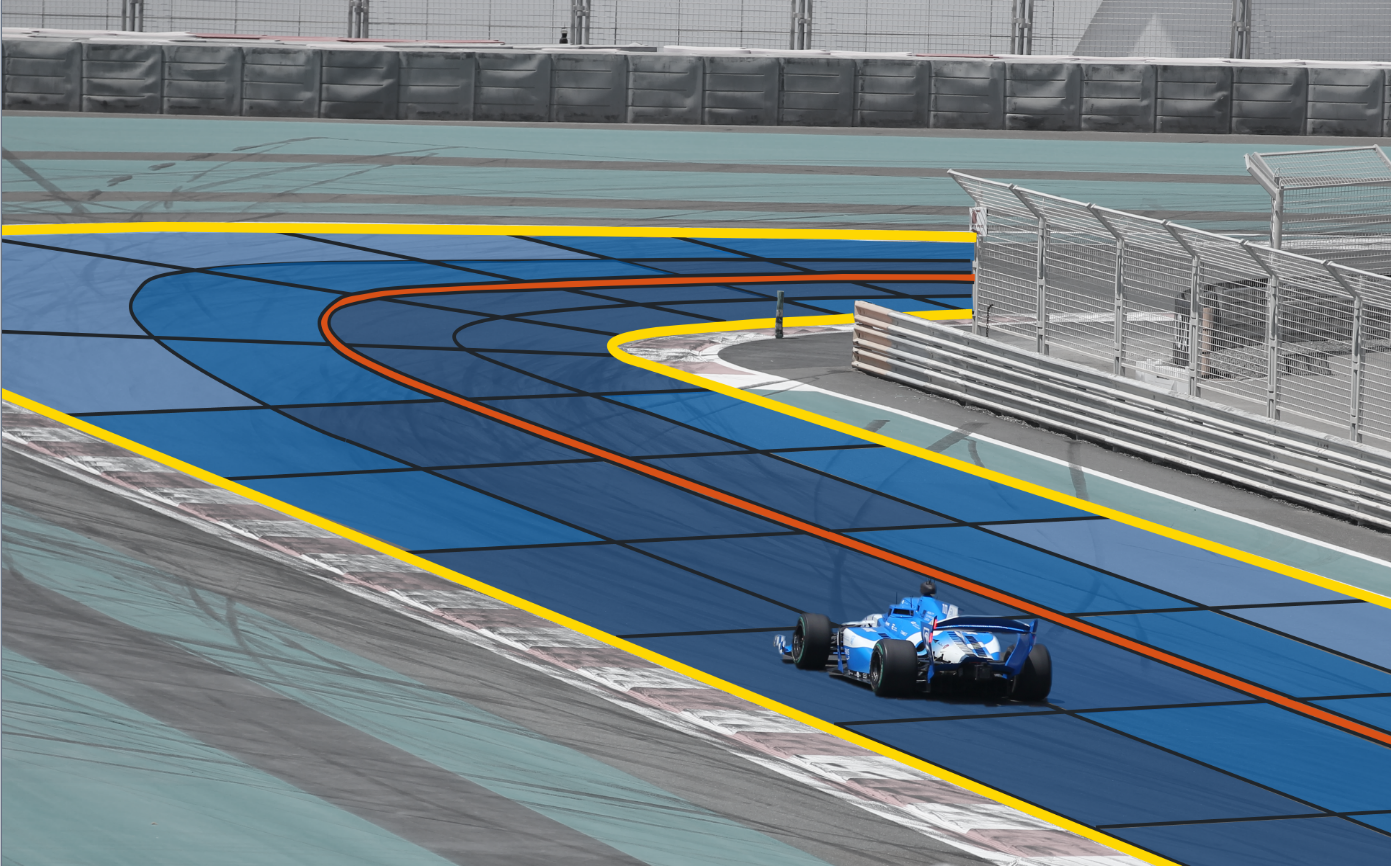}
\caption{GripMap structure in the Frenet frame, which is centered around a reference line (centerline).
The GripMap cells are discretized in the Frenet frame with constant step widths $\Delta s$ and $\Delta n$.
}
\label{fig:GMStruktur}
\end{figure}
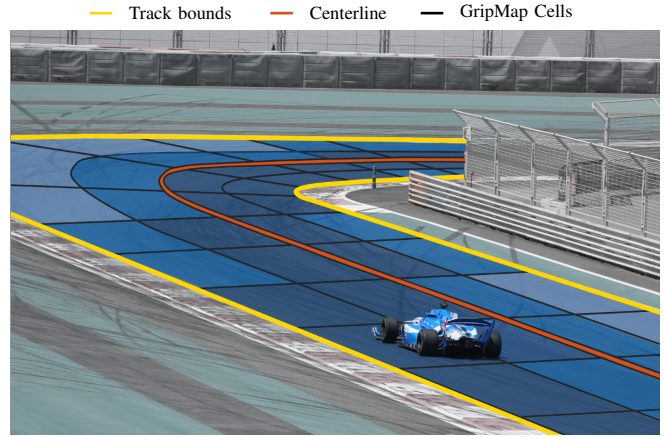

In this work, we propose \textit{GripMap}, a novel framework for spatially resolved vehicle dynamics constraints, which accounts for location-specific grip variations and software specific constraints.  Unlike conventional global friction models or robust methods that rely on worst-case assumptions, GripMap provides planning algorithms with location-dependent information about the usable dynamic potential of the entire vehicle system. This allows the planner to increase accelerations where the system, including physical grip and software capabilities, permits it, and to reduce them where constraints due to friction, control authority, modeling inaccuracies, or localization uncertainties demand caution. We discretize the track in the Frenet frame, storing local grip-scaling factors \(\theta_{ij}\) in a dense lookup structure for efficient real-time access. This paper highlights how GripMap is integrated into both a minimum lap time offline trajectory optimization and a high-frequency online planner, yielding faster lap times and improved reliability in real-world racing scenarios. 
A key emphasis is placed on efficient integration to minimize computational overhead. The GripMap framework targets the challenges experienced in real-world multi-vehicle racing scenarios. In summary, the paper has the following key contributions.

\begin{itemize}
    \item \textbf{Spatially Resolved Constraint Framework}
    We present GripMap, a novel method to resolve vehicle dynamics constraints in the Frenet frame, enabling local adaptation to varying grip conditions.
    \item \textbf{Efficient Perfect-Hash-Based Indexing}
    An indexing mechanism for a grid structure in Frenet frame is introduced, inspired by perfect hashing technique, ensuring minimal computational overhead and lightweight integration.
    \item \textbf{Simulation and Real-World Validation}
    We validate the effectiveness of the GripMap framework in real-world autonomous racing contexts. %, including the Indy Autonomous Challenge (IAC) and the Abu Dhabi Autonomous Racing League (A2RL). 
    Our experiments demonstrate improved lap times (by 5.2\% compared to spatially uniform models) and avoidance of grip overuse on surfaces with unknown or varying grip, thereby preventing accidents.
\end{itemize}

\section{RELATED WORK}

To be successful in autonomous racing~\cite{Betz.2022}, software needs to excel in two main aspects: Firstly, the ability to complete a lap in the fastest possible time ~\cite{Piccinini.2024b}. Secondly, the ability to react to other agents and maneuver around them effectively.
To achieve this, the current state-of-the-art approach splits the planning process into offline and online components~\cite{Betz.2023, Raji.2022, Jung.16.03.2023}. Offline, a Minimum Lap Time Problem (MLTP) is solved to generate a smooth and fast reference trajectory based on a given vehicle model and track geometry. A widely used method for solving the MLTP is the optimal control problem (OCP), which formulates a constrained optimization problem. In this formulation, the vehicle dynamics are represented as a set of differential equations, with time as the independent variable to be minimized~\cite{DalBianco.2019, Massaro.2021}.

Common approaches for modeling vehicle dynamics constraints in OCPs include the use of point-mass models with acceleration constraints~\cite{Brayshaw.2005, Veneri.2020, Lovato.2022, Rowold., Piccinini_ggv.2024} or single-track/double-track models incorporating tire dynamics~\cite{Tavernini.2013, Berntorp.2014, Perantoni.2014, DalBianco.2018, Gabiccini.2021}. However, these methods assume spatially invariant vehicle dynamics. The only publication that explicitly considers location-dependent vehicle models in MLTP OCPs is Christ et al.~\cite{Christ.2021}. Here, a time-MLTP OCP for race cars is introduced, that accounts for locally varying tire-road friction via a friction map. While they demonstrate how friction changes can significantly alter the racing line and lap times, the approach is purely offline and computationally demanding. Consequently, the method does not extend to high-frequency online re-planning in multi-vehicle racing scenarios.

Beyond explicit friction maps, several works have tackled \emph{time-varying friction} from a \emph{model-adaptive} perspective without spatial resolution. For instance, Svensson et al.~\cite{Svensson.2021,Svensson.2022} fuse slip-based friction with a vision-based surface classifier, but remain limited to a 1D friction estimate. Also their method is not designed to take advantage of the lap-by-lap repetitive nature of racetrack driving.

Similarly, Nagy et al.~\cite{Nagy.2023} and Kalaria et al.~\cite{Kalaria.3142023} address multi-friction adaptation by \emph{learning a time-varying vehicle model}. In~\cite{Nagy.2023}, a library of Gaussian Processes (GPs) captures dynamics under different friction conditions and is combined in real-time using a convex weighting scheme. Kalaria et al.~\cite{Kalaria.3142023} use Extreme Learning Machines (ELMs) to approximate the tire-slip curve online, enabling continuous adaptation over time. Both approaches, however, focus on the \emph{temporal} or operating-condition changes in friction rather than representing location-specific grip variability. As a result, while these methods can respond quickly to unexpected grip changes, they do not leverage repeated traversal or offline mapping to build a position-dependent friction model of the track.

To successfully address the challenge of online trajectory planning required for multi-vehicle racing, an algorithm  is required that can handle the non-convexity of multi-vehicle interactions while running fast enough to enable dynamic re-planning multiple times per second. To achieve consistency between the offline and online planning algorithms, the developed GripMap framework should be applicable in both algorithms and must, therefore, be numerically efficient to minimize impact on the online planning frequency.

\section{METHOD}
The proposed GripMap framework spatially resolves vehicle dynamic constraints by storing vehicle-model parameters in discrete grid cells of a Frenet based grid. In our baseline setup, we employ a point-mass model constrained by g-g-g-v limits, as introduced by Rowold et al.~\cite{Rowold.}, where lateral and longitudinal acceleration limitations are treated as dependent variables based on apparent vertical acceleration and velocity. 
We extend this model by introducing a dimensionless scaling factor \(\theta\) that is locally defined via the GripMap. This factor modulates the effective acceleration limit constraints for each discrete cell in the map. Although demonstrated with a point-mass model, the GripMap concept can be extended to single-track or double-track vehicle models by storing position-dependent tire parameters (e.g., \(\mu\) for the front and rear axles) in each cell. In the presented form, GripMap serves as an offline framework and does not adapt in real-time. It rather provides a systematic and efficient approach for integrating position-specific adaptations into vehicle dynamics constraints.

\begin{figure}
\centering
\includegraphics[width=0.95\columnwidth]{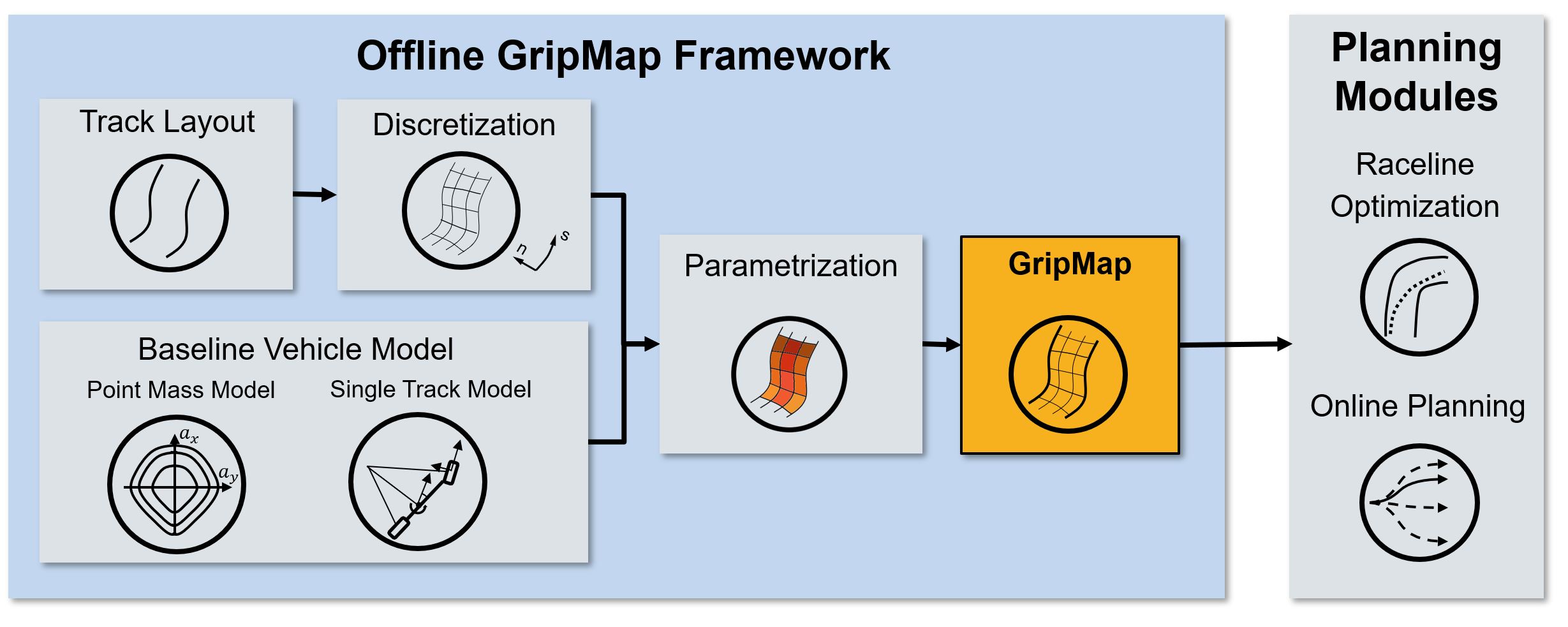}
\caption{Overview of our GripMap Framework: We discretize the track geometry in the Frenet frame and combine it with a baseline vehicle model. Each GripMap cell locally refines this model, in our case by applying a scaling factor $\theta$ to a point-mass model governed by g-g-g-v constraints. Together, these elements form the GripMap, which serves as the vehicle-dynamics constraint in our planning modules.
}
\label{fig:GMStruktur}
\end{figure}

The integration of GripMap into our autonomous driving planning algorithms~\cite{betz2023tum, Rowold., Oegretmen.2024} can be summarized as follows:  
First, an offline MLT problem is solved, where the vehicle’s trajectory is optimized to minimize overall lap time under incorporation of the GripMap’s local friction-scaling factors.  
During race operation, a high-frequency sampling-based planner continuously re-plans around the offline solution. This local planner queries the GripMap to check the feasibility of the targeted acceleration based on the local grip availability. If a candidate trajectory exceeds the local acceleration limit, it either incurs a penalty in the cost function or is rejected outright.

\subsection{GripMap Structure}\label{subsection:Struktur}
We discretize the track in the Frenet frame \((s, n)\), which is aligned with the track centerline and serves as the reference line. The Frenet frame defines a moving coordinate system along a reference curve, where \(s\) represents the arc length along the track centerline (longitudinal position) and \(n\) denotes the lateral offset perpendicular to the reference line. A basic representation of this GripMap structure is shown in Figure \ref{fig:GMStruktur}. The Frenet-based discretization is advantageous for both memory efficiency and computational speed, as the drivable area on a racetrack is typically concentrated near the centerline. By avoiding the discretization of large off-track regions, we reduce memory usage. As an example, discretizing the Yas Marina Circuit in Abu Dhabi in the Frenet frame saves ~90\% memory compared to Cartesian.

Let \(s_{\max}\) be the total arc length of the centerline of of the track. We divide this arc length into \(M = s_{\dim}\) equal-length segments:
\begin{equation}
  \Delta s = \frac{s_{\max}}{s_{\dim}},
\end{equation}
resulting in discrete longitudinal positions \(s_i = i \cdot \Delta s\) for \(i \in \{1, 2, \dots, s_{\dim}\}\) denoting the upper limit in s direction of the respective cells.

For the lateral coordinate \(n\), we introduce a discretization parameter \(n_{\dim}\) to define the resolution. First, we determine the maximum lateral extent of the racetrack, denoted as \(W_{\max}\). We then compute the lateral step size:
\begin{equation}
  \Delta n = \frac{W_{\max}}{n_{\dim}}.
\end{equation}
Using this step size, we construct a symmetric lateral grid around \(n=0\) with an equal number of grid cells on each side. To achieve symmetry, we generate \(n_{\dim}\) positive and \(n_{\dim}-1\) negative segments. Each value in the set represents the left/lower boundary of its respective cell. The lateral coordinates are thus defined as:

\begin{equation}
  \mathbf{n} = 
  \begin{aligned}
    &\{-n_{\dim} \cdot \Delta n, \, -(n_{\dim}-1) \cdot \Delta n, \, \dots, \, -\Delta n, \, 0, \, \\
    &\Delta n, \, \dots, \, (n_{\dim}-1) \cdot \Delta n\}
  \end{aligned}
\end{equation}

This results in \(2n_{\dim} = N\) discrete lateral positions \(n_j\) for \(j \in \{-n_{\dim}, \dots, 0, \dots, n_{\dim}-1\}\) denoting the lower boundary in n direction of the respective cells.

Combining the longitudinal and lateral discretization, we form an \(M \times N\) grid, where each cell \((s_i, n_j)\) stores a local scaling factor \(\theta_{ij}\). In a point-mass model constrained by g-g-g-v limits, let
\begin{equation}
  a_{x,\mathrm{max,base}}(v, a_z), \quad a_{y,\mathrm{max,base}}(v, a_z)
\end{equation}
denote the baseline maximum longitudinal and lateral accelerations at velocity \(v\) and vertical acceleration \(a_z\). The effective local constraints are then given by:
\begin{equation}
  a_{x,\mathrm{max}}(s_i, n_j) = \theta_{ij} \cdot a_{x,\mathrm{max,base}}(v, a_z),
\end{equation}
\begin{equation}
  a_{y,\mathrm{max}}(s_i, n_j) = \theta_{ij} \cdot a_{y,\mathrm{max,base}}(v, a_z).
\end{equation}

\subsection{Indexing and Lookup}
By leveraging the Frenet frame for discretization, GripMap is implemented as a dense two-dimensional lookup matrix. To find the index of the corresponding grid cell \((s_i, d_j)\) for a continuous Frenet coordinate \((s, d)\), we first compute the discrete longitudinal index \(i\) by division of \(s\) by \(\Delta s\):
\begin{equation}
  i = \left\lfloor \frac{s}{\Delta s} \right\rfloor,
\end{equation}
where \(\lfloor \cdot \rfloor\) denotes the floor operation. We clip \(i\) to ensure \(0 \leq i < s_{\dim}\). For the lateral coordinate, we introduce an offset \(n_\text{offset}\) so that \(n = 0\) aligns with a central index. The discrete index \(j\) is calculated by
\begin{equation}
  j = \left\lfloor \frac{n + n_\text{offset}}{\Delta n} \right\rfloor,
\end{equation}
and is clipped to \(0 \leq j < N\). Once \((i, j)\) is obtained, the local scaling factor \(\theta_{ij}\) (or any other stored parameter, such as \(\mu_{ij}\) in a single-track model) is retrieved from the \(M \times N\) matrix. This perfect hashing approach enables GripMap queries in \(O(1)\) time. Additionally, the contiguous storage of the GripMap enhances memory efficiency by avoiding pointer and traversal overhead.

\subsection{GripMap Integration into Offline Raceline Optimization}
The GripMap is used in the computation of the offline raceline. The offline raceline provides a trajectory that serves as a reference for the sampling-based online planner. The MLTP is solved by formulating an OCP following the approach in~\cite{Rowold.}. The track is represented as ribbon-based in three dimensions, allowing for the depiction of banked turns, as well as dips and crests. As a vehicle model, we use a point mass model constrained by g-g-g-v diagrams. We model the raceline as a sequence of quasi-steady states, regularizing the gradient between consecutive states to ensure smooth transitions. 

The constraints for the OCP are based on those formulated by~\cite{Rowold.}.
Equation~(\ref{eq:OCP}) outlines the OCP along with its key constraints, now employing g-g-g-v limits extended by a local scaling factor $\theta_{ij}$ from the GripMap:
\begin{equation}
\label{eq:OCP}
\begin{aligned}
\min_{x,u} \!\!
& \int_{s_0}^{s_e}\!\Bigl(
    \tfrac{1}{\dot{s}} 
    + u^T R\,u 
\Bigr)\,ds \\[4pt]
\text{s.t.}\quad
& \frac{d\textbf{x}}{ds} = \frac{1}{\dot{s}}\,\textbf{f}(\textbf{x},\textbf{u}), \\[3pt]
& g\bigl(V,a_z,a_x,a_y,\theta_{ij}\bigr)\,\le\,0, 
    \quad (\text{ g-g-g-v constraints}),\\[3pt]
& n_{\min}\le n \le n_{\max},\;\; 
  -\tfrac{\pi}{2}\le\hat{\chi}\le\tfrac{\pi}{2}, \\
& \textbf{x}(s_0 = 0) = \textbf{x}(s_e = s_f)\;
\end{aligned}
\end{equation}

Where the terms are defined as follows:
\( s \) represents the arc length, with \( \dot{s} = ds/dt \).  
\( x(s) \) denotes the vehicle state, which includes parameters such as \( V \), \( n \), \( \hat{\chi} \), \( a_x \), and \( a_y \).  
The input vector \( u(s) = [\hat{\,j_x\,}\; \hat{\,j_y\,}]^T \) contains the longitudinal and lateral jerk.  
\( R \) is the jerk regularization.  
\( \theta_{ij} \) is the dimensionless GripMap scale factor at \( (s_i, n_j) \).  
\( g(\cdot) \leq 0 \) refers to the modified acceleration constraints defined via g-g-g-v limits and \( \theta_{ij} \).  
\( n \) is the lateral deviation from the reference line.  
Finally, \( \hat{\chi} \) represents the relative orientation to the reference line.

\subsection{GripMap Integration into the Sampling-Based Online Planner}
We employ a high-frequency sampling planner based on~\cite{Oegretmen.2024}, which generates multiple candidate trajectories from the current Frenet state \(\bigl(s_0, \dot{s}_0, n_0, \dot{n}_0\bigr)\) and selects one for execution based on a cost metric. The GripMap integration occurs during the cost evaluation step, where sampled trajectories are assessed against local g-g-g-v constraints retrieved from the GripMap. Compared to the original approach, we shift the acceleration checks from a strict feasibility filter to a soft-constraint term in the cost function, allowing limited grip overuse balanced against other cost terms. The algorithm involves three steps: 

\begin{enumerate}
    \item \textbf{Sampling and Feasibility Checks}  
    The planner samples candidate end states \(\bigl(\dot{s}_e, n_e\bigr)\) and connects each to the initial state via polynomial trajectories. We perform track-boundary checks and ensure minimal turning radii are not exceeded. However, the local acceleration constraints are no longer enforced as hard limits.  

    \item \textbf{Cost Function Evaluation}  
    Trajectories that exceed the local g-g-g-v acceleration thresholds (scaled by \(\theta_{ij}\) from the GripMap) incur an additional cost rather than immediate rejection. This soft-penalty mechanism balances grip-limit exceedance against other factors such as opponent proximity or tracking the racing line. Consequently, the planning algorithm can briefly exceeds nominal limits if that lead to a reduction in collisions risk or leads to a strategic advantage in close racing scenarios.

    \item \textbf{Trajectory Selection and Re-Planning}  
    Each candidate trajectory’s total cost is computed by summing costs for all employed cost terms. The planner selects the lowest-cost path and restarts the entire sampling process with the updated ego and opponent vehicle states.
\end{enumerate}

\section{RESULTS}

\subsection{GripMap Parametrization and Iterative Adjustment}\label{subsec:iterative_adjustment}
Accurately parametrizing the GripMap poses a challenge, as the true friction distribution across a racetrack is rarely known a priori. Although full-scale track measurements can theoretically be conducted, as demonstrated by Waddell~\cite{Wadell.2019}, practical constraints on equipment and track access often preclude such exhaustive approaches. 

\begin{figure}
\centering
\includegraphics[width=1\columnwidth]{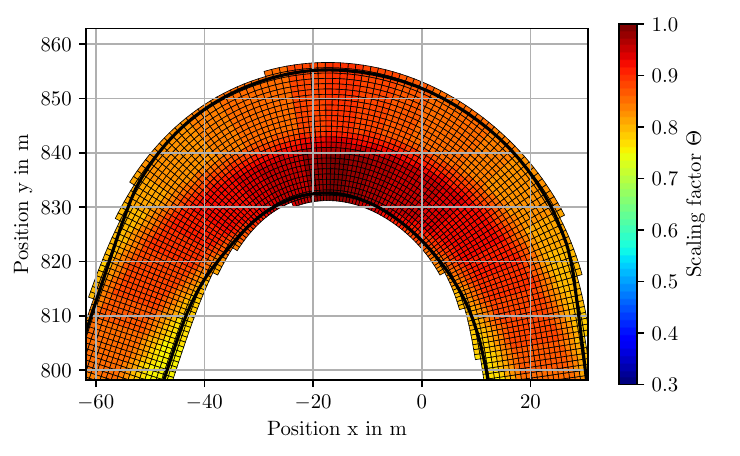}
\caption{GripMap discretization and corresponding scaling values for Turn 5 of the Yas Marina Circuit in Abu Dhabi. The highest grip is observed on the racing line, with decreasing grip values further away, reflecting the increased accumulation of debris and dirt on less frequently driven areas of the track.}
\label{fig:GMT5}
\end{figure}

Instead, we followed an iterative process in our participation in last year’s A2RL competition. The A2RL competition is an autonomous multi-vehicle racing competition taking place on the Yas Marina Circuit in Abu Dhabi. Multiple challenges were held, including qualification runs, head-to-head races, and a four-vehicle final race.
Initializing the GripMap with a conservative baseline, we incrementally increased the local scaling factors whenever post-run data analysis revealed that the tires were still underutilized and the localization and control modules demonstrated reliable and desired behavior. While the specific data analysis techniques and validation steps used in each iteration varied and are not detailed in this work, the goal was to demonstrate that such an iterative refinement process is feasible in real-world scenarios. In autonomous racing, the software acts as the driver and often limits vehicle performance. GripMap captures not only the physical grip potential but also the combined capabilities of the \textit{vehicle-driver system}. If control errors or oscillations repeatedly occur in a specific section-indicating a risk of instability-local scaling factors \(\theta_{ij}\) are reduced to reflect these software constraints. This adjustment maintains safety, even if a more aggressive trajectory might be feasible with a different software stack. Through iterative data collection, updates, and on-track validation, we progressively refined GripMap. Interpreting \(\theta_{ij}\) as a constraint on the full system and not only on friction potential, enables reliable performance under real-world limitations.

\begin{figure}
\centering
\includegraphics[width=0.95\columnwidth]{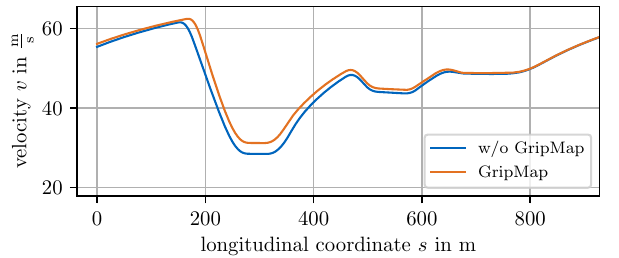}
\caption{Velocity profiles for an Abu Dhabi lap obtained via offline raceline optimization using a uniform 75\% global scaling factor versus the final spatially resolved GripMap (ranging between 75\% and 100\%). The locally varying scaling factors lead to noticeable differences in braking points and corner speeds, highlighting the impact of utilizing spatially variant dynamic constraints.}
\label{fig:velocity comparison}
\end{figure}

This iterative approach proved highly beneficial in the A2RL competition. With multiple iterative refinements, we adjusted the GripMap’s scaling factors to exploit our vehicle system’s full potential with the resulting velocity profile shown in Figure \ref{fig:velocity comparison}. The resulting values ranged from \(75\%\) to \(100\%\) of our baseline g-g-g-v estimate, demonstrating a wide spread of grip availability and system performance across the track. Figure \ref{fig:GMT5} displays a section of the final GripMap iteration used in A2RL. Without the flexibility of a spatially resolved GripMap, we would have been forced to adopt the worst case global scaling factor of \(75\%\) across the entire track to ensure feasibility in the most limiting corner. With GripMap, we maintained this conservative value only in that specific region while using higher scaling factors elsewhere — leading to an overall lap-time improvement of \(5.2\%\).

\subsection{Runtime Evaluation}
\label{subsec:runtime_eval}

To evaluate the runtime overhead introduced by the GripMap lookup in the acceleration checks as described in Section \ref{subsection:Struktur}, we compare the average runtime of the GripMap informed online planning module variant against a baseline without GripMap. Both configurations are implemented as ROS2 Python nodes based on Oegretmen et al.~\cite{Oegretmen.2024} and executed on a machine equipped with an Intel Core i7-11850H CPU. To ensure consistent performance measurements, the planning node is pinned to an isolated core with hyperthreading disabled.
Each configuration was run three times, with each run lasting fifteen minutes in a qualification lap configuration.
In the “GripMap informed” configuration, a lookup by the Frenet coordinate $(s, n)$ is performed for every discretization point on each sampled trajectory that falls within the track bounds. In this experiment, each planning step generated $1{,}000$ sampled trajectories and each trajectory is discretized into $40$ points.

\begin{table}[H]
    \centering
    \caption{Planner Runtime With vs.\ Without GripMap}
    \label{tab:runtime_comp}
    \begin{tabular}{l
                    S[table-format=1.9]
                    S[table-format=1.9]
                    S[table-format=5.0]}
        \toprule
        \textbf{Configuration} & {\textbf{Mean} [s]} & {\textbf{Std. Dev.} [s]} & {\textbf{Samples}}\\
        \midrule
        GripMap Informed & 0.1565 & 0.0280 & 17678\\
        No GripMap & 0.1553 & 0.0303 & 18372\\
        \bottomrule
    \end{tabular}
\end{table}

\noindent
As shown in Table~\ref{tab:runtime_comp}, the difference between these two mean runtimes is about $0.0012\,\mathrm{s}$ (or $\sim 0.77\%$). Overall, the additional overhead of the GripMap lookup is minor in both absolute and percentage terms. Given the benefits of location-resolved grip constraints, we consider this runtime penalty acceptable for our application.

\subsection{Safe Multi-Vehicle Interactions in Varying Grip Conditions}
\label{sec:multi_vehicle_interactions}

Autonomous racing encounters significant challenges during multi-vehicle interactions, particularly when maneuvers like overtaking must be performed off the racing line where less data is available. While many online planners assume globally invariant vehicle dynamic models across the entire track, real-world race track conditions can vary significantly. Especially when deviating laterally from the common racing line, the available grip can differ substantially and is often known with lower certainty due to a lack of prior data. 

\begin{figure}
\centering
\includegraphics[width=1\columnwidth]{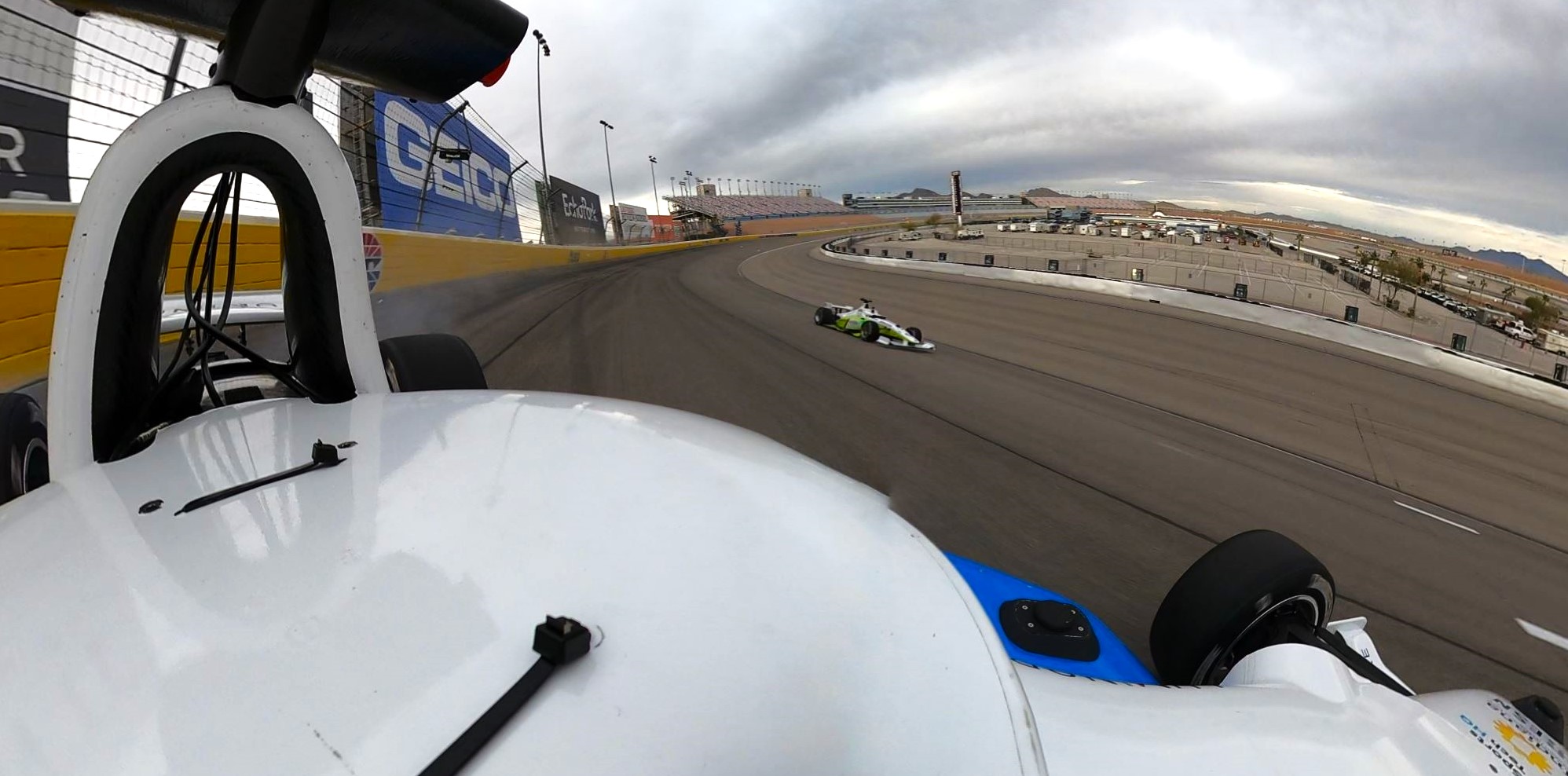}
\caption{Beginning of a spin-out during the IAC 2023 final at Las Vegas Motor Speedway. Dust kicked up by the tires on the outside line of a banked turn is clearly visible. This dust on the driving surface reduced grip availability, and without a spatial resolution of the vehicle dynamic constraints, the planning algorithm generated a dynamically unfeasible trajectory. The control algorithm, attempting to follow this unfeasible trajectory, ultimately led to the observed spin.}
\label{fig:Spin}
\end{figure}

A demonstrative example occurred in the 2023 \emph{Indy Autonomous Challenge~(IAC)} finals at the Las Vegas Motor Speedway. The TUM Autonomous Motorsport vehicle was attempting an overtake against PoliMOVE in a high-speed banked turn during an autonomous multi-vehicle challenge. The attempt ended in a spin because the software planned its trajectory on a section of the track where dirt accumulation led to a reduction of the available grip. The beginning of the spin is shown in Figure \ref{fig:Spin}. The PoliMOVE vehicle was constrained by a rule-based speed cap of \SI{155}{mph} (approximately \SI{69}{m\per s}), which they temporarily had to adhere to. The TUM vehicle was not under the restriction of a speed cap and was free to reach the maximum speed of the vehicle of approximately \SI{78}{m\per s}, in order to gain a speed surplus to initiate the overtake maneuver. Since PoliMOVE chose to defend on an inside line in close proximity to the bottom of the elevated banking turn, the remaining option was to initiate a pass around the outside. Footage of the overtake attempt is available: https://www.youtube.com/watch?v=wZfNukyiciY.

\begin{figure}
    \centering
    \resizebox{0.95\columnwidth}{!}{\input{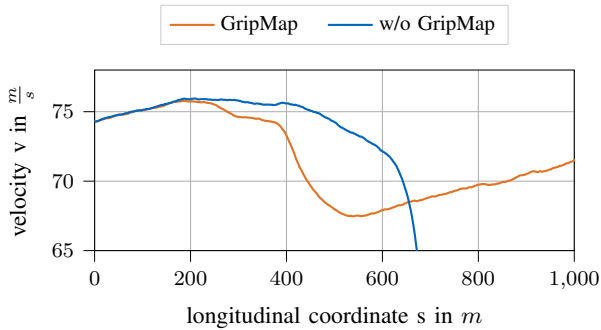}} 
    \caption{Comparison of ego velocity profiles during an outside overtaking maneuver at the Las Vegas Motor Speedway. The globally invariant vehicle model without a GripMap overestimates grip at the far outside line, whereas a GripMap informed planning module reduces the corner entry velocity by up to $7\frac{m}{s}$ where the vehicle enters the area of reduced grip.}
    \label{fig:velocity_profiles}
\end{figure}

At the time, no spatial-resolved grip information was available to the sampling-based online planner and in this situation, the online planner’s cost function was balancing several main factors which influenced the chosen trajectory. In this specific instance the evaluation of the cost terms leaned more toward maintaining distance from the opponent and reducing the lateral acceleration jerk than staying close to the racing line. As a result, the planner chose a trajectory that involved moving to a higher lateral position on the banked turn. Given the absence of spatially resolved vehicle dynamic constraints, the sampling-based online planner selected a fast trajectory that moved to the outside of the turn. While the trajectory appeared dynamically feasible within a globally invariant vehicle model, it failed to account for the accumulation of dust and debris on this track section. The dust in the air behind the vehicle can be observed in Fig.~\ref{fig:Spin}. Although prior runs confirmed that the targeted accelerations were achievable on the racing line, data analysis revealed that the available grip on the chosen trajectory used for the overtake attempt was approximately \SI{35}{\percent} lower than assumed. The requested trajectory led to a loss of vehicle control, causing a spin which ended the race.

This incident serves as the prime motivation for applying the GripMap framework. To further investigate the situation, we incorporate a \emph{ground truth GripMap} within a validated high-fidelity vehicle-dynamics simulation~\cite{Sagmeister.2024} to replicate these conditions by lowering friction scaling off-line in the vehicle-dynamics simulation. Specifically, we adjust the lateral and longitudinal friction scaling factors (\texttt{LMUY}, \texttt{LMUX}) of the utilized Pacejka MF52 tire model depending on the lateral coordinate~$n$. We define a plateau of maximum grip near the racing line and reduce \texttt{LMUY} and \texttt{LMUX} linearly with increasing distance from it, replicating the experienced real-world conditions. By recreating the overtake scenario in simulation with this ground truth grip map, we can replicate the observed spin. Conversely, when the online planner is informed by the GripMap, it avoids exceeding local friction limits, reduces corner entry speed and the planned trajectory moves further downward on the banked turn to a section with higher grip availability. The planned tire utilization (percentage of the assumed g-g-g-v usage) in this instance never exceeds 100\%.

\begin{figure}
    \centering
    \resizebox{0.95\columnwidth}{!}{\input{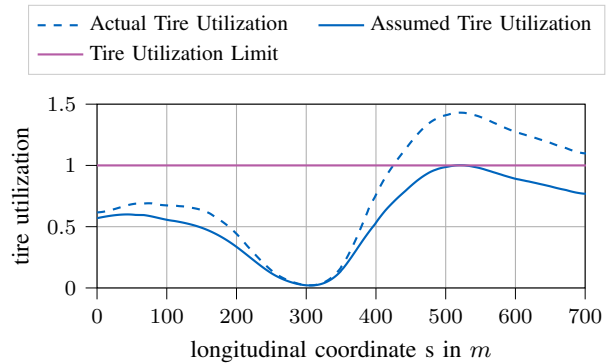}} 
    \caption{Tire utilization in one overtaking scenario. The solid line indicates the assumed tire utilization without considering local grip changes. In contrast, the dashed line shows what the actual tire utilization was, using the presumed track grip. }
    \label{fig:tire_util}
\end{figure}

\begin{figure*}[ht]
    \centering
    \resizebox{0.8\textwidth}{!}{\input{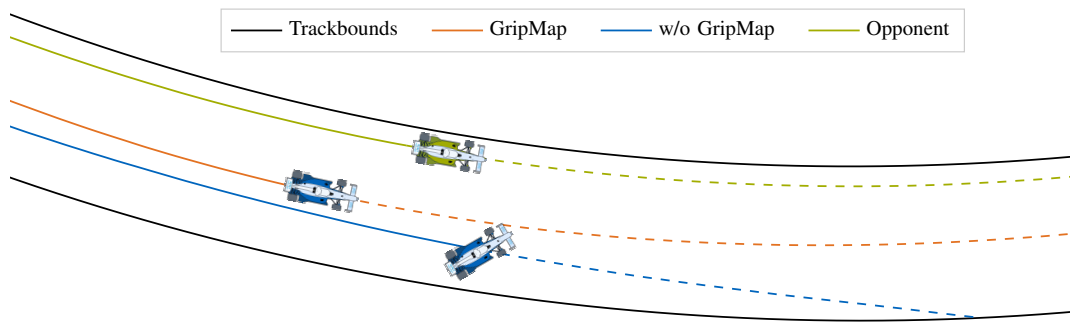}} 
    \caption{Resimulation of the interaction situation with and without GripMap Information. The positions of the vehicles are shown at approximately the same time. PoliMOVE (green vehicle) stays on an inner line and is on a similar position in both simulations, while the TUM cars show different behavior. The blue car with the blue trajectory lacks GripMap information, while the GripMap informed vehicle (blue car/orange trajectory) vehicle enters the turn at a reduced velocity and on a more inside lateral position offering increased grip.}
    \label{fig:positions}
\end{figure*}

Figure~\ref{fig:velocity_profiles} compares the velocity profiles. Without GripMap information, the planner overestimates the grip off the main racing line, leading to target lateral accelerations not achievable after extending the simulator with the \emph{ground truth GripMap}. Figure~\ref{fig:tire_util} displays the tire utilization of the simulation run without the incorporation of the GripMap. The assumed tire localization shows that the planner was utilizing the assumed grip potential available and operating correctly within the set bounds. However, if we take the \emph{ground truth GripMap} into account, the actual tire utilization reached values of over \SI{140}{\percent}, leading to a similar outcome to what has been observed in the real-world example. The driven trajectories in this scenario are illustrated in Figure \ref{fig:positions}. While PoliMOVE is depicted in the same position, the TUM car utilizing the GripMap is still behind PoliMOVE due to its lower turn-entry speed and laterally closer to the raceline due to the grip information. In contrast, the TUM car without GripMap-informed planning already starts spinning due to an overestimation of the available grip.

\section{DISCUSSION}
In this section, we compare the proposed GripMap framework and alternative strategies for addressing unknown or varying grip conditions in autonomous racing. Additionally, we analyze current challenges and further improvement possibilities for the GripMap. 

\subsection{Key Findings}
We hypothesized that spatially resolved vehicle dynamic constraints would allow exploitation of higher friction where available while remaining conservative in regions with lower or uncertain grip. The 5.2\% lap-time improvement on the Yas Marina Circuit supports this view. While a full statistical analysis was not feasible due to the empirical nature of the iterative tuning, the relative performance benefit over the baseline global model is consistent across iterations and supported by observed trajectory adaptations.

The spin-out incident at the Las Vegas Motor Speedway underscores the limitations of global vehicle models in practice. Although time-varying or model-adaptive approaches can capture large-scale changes, such as shifting weather or tire wear, they often lack reliable means to extrapolate grip to uncharted regions or to vary it consistently across the track. In high-speed racing, overtaking or defending can force a car onto areas where localized grip differs significantly. Global friction assumptions can lead to the generation of dynamically unfeasible trajectories, which led to the spin in Las Vegas. Our simulation recreated the spin to confirm that incorporating spatially resolved constraints would have averted this scenario. Thus, our results reinforce the initial hypothesis that spatially resolved vehicle models yields both performance gains and safer race maneuvers.

\subsection{Parametrization Challenges and Data Gathering}
As highlighted in the literature~\cite{Woodward.2012, Wadell.2019} and demonstrated in section~\ref{subsec:iterative_adjustment} a race track can show considerable variability in available grip especially when leaving the racing line. However, full spatial grip estimates for a given race track are rarely available and subject to change in between sessions on this track, tire and setup changes or through weather events. Even though we have shown promising results in the A2RL competition in 2024, multiple consecutive iterations were necessary to adjust the GripMap for this specific software-vehicle-track combination. Additionally the GripMap parametrization was done on the race line only, extrapolating to the areas besides the raceline based on empirical heuristics, because of time and data availability constraints. Therefore, the transferability to other tracks and track conditions as well as the accuracy in regions off the frequently driven racing line remains limited. A formalized method for initializing and refining GripMap scaling factors is not covered in this paper, as the focus lies on the framework itself and its integration. Future work will address this shortcoming through automated or learning-based parametrization techniques.

\subsection{Vehicle versus Driver Constraints}
A notable benefit of the GripMap is its ability to capture both local friction limitations and software-specific limitations using the vehicle dynamic constraints. While prior research focuses on tire-road limitations~\cite{Svensson.2021,Svensson.2022,Nagy.2023,Kalaria.3142023}, the total vehicle system potential is also limited by the software capability. This is analogous to human drivers, as our previous study has shown, the driver can in fact always be viewed as the limiting factor and in the best case the vehicle dynamic limit is congruent with the driver skill limit~\cite{Werner.2024}. The current GripMap design considers both, but it cannot be distinguished as it is both lumped into the scaling factor \(\theta_{ij}\). While this simplifies real-world deployment, it can obscure whether local friction conditions or the control algorithm limit performance. Separating these effects in the future may be better suited to generalize to different software or evolving environmental conditions.

\subsection{Toward Real-Time Adaptation}
The GripMap concept has the inherent advantage of interpretability: each cell corresponds to a measurable or learned observation at a particular track location. This makes it well-suited as a foundation for explainable, learning-based approaches that can update the GripMap during runtime. However, the current static \emph{GripMap} design lacks adaptability to dynamic conditions, like evolving rubber deposits, tire temperatures or weather conditions. Still, by demonstrating real-time compatibility of spatially resolved vehicle dynamics, this work lays the foundation for future learning algorithms that update the GripMap during runtime, without reengineering the trajectory planning integration.

\section{CONCLUSION AND OUTLOOK}
This work presented the GripMap framework, which enables spatial resolution of vehicle dynamic constraints in both offline and online planning algorithms for autonomous racing. By leveraging a Frenet-based grid and perfect-hash-based indexing, GripMap supports memory-efficient data storage and constant-time lookups, enabling real-time trajectory planning with spatially resolved constraints, inducing only 0.77\% of additional runtime in a real-time execution experiment. While the GripMap itself remains static during runtime, this evaluation demonstrates that incorporating spatially resolved vehicle models into real-time planning is computationally feasible, addressing a key limitation of prior work~\cite{Christ.2021}. Experimental results demonstrate that locally adapted grip constraints can improve lap times and enable safe multi-vehicle interaction on unexplored surfaces, thus enhancing both performance and reliability in high-speed racing environments. The incorporation of the GripMap into the planning algorithm's vehicle dynamic constraints ensures that planned trajectories remain dynamically feasible, and therefore enhances the robustness of multi-vehicle interaction maneuvers against real-world surface variability. This mitigates the risk of loss of vehicle control and has broader relevance beyond motorsport, for example for highway driving where lanes may exhibit different road-friction characteristics due to weather, wear, or debris. 

Future work will focus on integrating learning-based methods so that a software stack is able to self-identify the GripMap online, accounting for factors such as evolving track and tire conditions or localized control constraints. Because the GripMap is interpretable and has proven its integrability into existing offline and online planning modules, it provides a robust foundation for more advanced, adaptive algorithms in autonomous motorsports and potentially broader automotive applications.

\addtolength{\textheight}{-12cm}   % This command serves to balance the column lengths
                                  % on the last page of the document manually. It shortens
                                  % the textheight of the last page by a suitable amount.
                                  % This command does not take effect until the next page
                                  % so it should come on the page before the last. Make
                                  % sure that you do not shorten the textheight too much.

%%%%%%%%%%%%%%%%%%%%%%%%%%%%%%%%%%%%%%%%%%%%%%%%%%%%%%%%%%%%%%%%%%%%%%%%%%%%%%%%

%%%%%%%%%%%%%%%%%%%%%%%%%%%%%%%%%%%%%%%%%%%%%%%%%%%%%%%%%%%%%%%%%%%%%%%%%%%%%%%%

%%%%%%%%%%%%%%%%%%%%%%%%%%%%%%%%%%%%%%%%%%%%%%%%%%%%%%%%%%%%%%%%%%%%%%%%%%%%%%%%
%\section*{APPENDIX}

%\section*{ACKNOWLEDGMENT}

\bibliographystyle{IEEEtran}
\bibliography{Literatur_Frederik_Werner_TUM_File}

\end{document}